\documentclass{article}

\usepackage{arxiv}

\usepackage[utf8]{inputenc} 
\usepackage[T1]{fontenc}    
\usepackage{hyperref}       
\usepackage{url}            
\usepackage{booktabs}       
\usepackage{amsfonts}       
\usepackage{nicefrac}       
\usepackage{microtype}      
\usepackage{lipsum}
\usepackage{graphicx}

\usepackage{algpseudocode}
\usepackage{lipsum}
\usepackage{fancyvrb}
\usepackage{parcolumns}
\usepackage{verbatim}
\usepackage{xcolor}
\usepackage{listings}
\usepackage{bera}

\usepackage{tikz}
\usetikzlibrary{shapes.geometric, arrows}
\tikzstyle{startstop} = [rectangle, rounded corners, minimum width=3cm, minimum height=1cm,text centered, draw=black, fill=red!30]
\tikzstyle{io} = [trapezium, trapezium left angle=70, trapezium right angle=110, minimum width=3cm, minimum height=1cm, text centered, draw=black, fill=blue!30]
\tikzstyle{process} = [rectangle, minimum width=3cm, minimum height=1cm, text centered, draw=black, fill=orange!30]
\tikzstyle{process2} = [rectangle, minimum width=2cm, minimum height=1cm, text centered, text width=2cm, draw=black, fill=red!30]
\tikzstyle{process3} = [rectangle, minimum width=3cm, minimum height=1cm, text centered, text width=3cm, draw=black, fill=blue!30]
\tikzstyle{decision} = [diamond, minimum width=3cm, minimum height=1cm, text centered, draw=black, fill=green!30]
\tikzstyle{arrow} = [thick,->,>=stealth]
\tikzstyle{line} = [thick,-,=stealth]

\title{Training distributed deep recurrent neural networks with mixed precision on GPU clusters.}

\author{
  Alexey Svyatkovskiy \\
  Princeton University\\
  Princeton, NJ 08540 \\
  \texttt{alexeys@princeton.edu} \\
   \And
Julian Kates-Harbeck \\
  Harvard University\\
  Cambridge, MA 02138 \\
  \texttt{jkatesharbeck@g.harvard.edu}
   \And
  William Tang \\
  Princeton University\\
  Princeton, NJ 08540 \\
  \texttt{wtang@princeton.edu}
}

\begin{document}
\maketitle

\begin{abstract}
In this paper, we evaluate training of deep recurrent neural networks with half-precision floats. We implement a distributed, data-parallel, synchronous training algorithm by integrating TensorFlow and CUDA-aware MPI to enable execution across multiple GPU nodes and making use of high-speed interconnects. We introduce a learning rate schedule facilitating neural network convergence at up to $O(100)$ workers. 

Strong scaling tests performed on clusters of NVIDIA Pascal P100 GPUs show linear runtime and logarithmic communication time scaling for both single and mixed precision training modes. Performance is evaluated on a scientific dataset taken from the Joint European Torus (JET) tokamak, containing multi-modal time series of sensory measurements leading up to deleterious events called plasma disruptions, and the benchmark Large Movie Review Dataset~\cite{imdb}. Half-precision significantly reduces memory and network bandwidth, allowing training of state-of-the-art models with over 70 million trainable parameters while achieving a comparable test set performance as single precision.
\end{abstract}


\section{Introduction}

Training deep neural networks is a computationally intensive problem which requires engagement of high-performance computing (HPC) clusters. Several implementations of distributed Stochastic Gradient Descent (SGD) exist, exploiting general-purpose hardware (CPU~\cite{cpu} and GPU~\cite{gpu}) as well as FPGA~\cite{fpga}. 

Data-parallel distributed training algorithms keep a copy of an entire neural network model on each worker, processing different mini-batches of the training dataset on each in parallel lock step. Some advantages of using synchronous SGD implementations are a reliable model convergence, ease of debugging, and avoidance of stale gradients. When training state-of-the-art large models (with over $O(10^{7})$ trainable parameters~\cite{DBLP:journals/corr/JozefowiczVSSW16}) the net gradient size per iteration can reach up to a few GB, potentially exceeding the limits of the network data transfer rate (network bandwidth) or the GPU device memory.


It has been shown that deep convolutional and recurrent neural networks allow using lower precision at both the training and inference stages in application to image classification and language modeling tasks~\cite{courbariaux+al-TR2014,Lin:2016:FPQ:3045390.3045690,hubara+al-2016-quantized}. Nevertheless, no extensive studies have been performed for recurrent neural networks during distributed training, which are at the focus of this paper. 

Compared to single precision, using half-precision floats allows training models with twice as many parameters, improves the maximum computational throughput, and optimizes memory and network bandwidths. At the time of writing, half-precision computation is supported by GPUs only, while the latest generation of Intel CPUs (e.g. Haswell, Broadwell) provide a capability for converting between single and half floating point precisions in hardware by means of the F16C instruction set. 

A set of strong scaling tests is performed using a framework integrating TensorFlow~\cite{tensorflow2015-whitepaper} with custom parameter averaging and global weight update routines implemented with CUDA-aware MPI. The software stack makes use of CUDA 8, GPU accelerated deep learning primitives from CuDNN 6, and TensorFlow 1.3. The performance is reported on a scientific dataset taken from the JET tokamak -- the largest operating tokamak in the world~\cite{vega2013results} -- comprising over 4000 time series, some of which result in deleterious events called plasma disruptions. The time series contain multi-modal sensory measurements, including both scalars and one-dimensional arrays, collected with a sampling rate of 1 ms. The description of the JET dataset is provided in Appendix~\ref{appendix:jet}. 

In addition to the JET dataset, we repeat the calculation on the Large Movie Review Dataset (IMDB)~\cite{imdb}, which is a public dataset often used for sequence learning and time-series forecasting tasks in deep learning community.

Following~\cite{Julian}, our goal is to train the recurrent neural net to predict the onset of a disruption at least $30~$ms in advance of the actual event. This is the time that would be required to implement mitigation strategies (such as rapid neutral gas injection) in a real experiment after receiving an alarm. The nature of the problem of disruption forecasting in tokamak fusion plasmas makes long short-term memory (LSTM) networks a promising candidate. 

The paper is organized as follows: we start with the hardware specifications and an overview of recurrent neural networks in the Sections I and II; Section III describes the details of the distributed training algorithm implemented in the framework; Section IV evaluates training with half-precision floats; Section V summarizes the performance on the plasma disruption dataset from JET. Section VI concludes the paper.  

\section{Hardware specifications}

\begin{figure*}
\begin{center}
\begin{tikzpicture}[node distance=4cm]
\node (in1) [process3] {Intel OmniPath};
\node (cpu0) [process, right of = in1, yshift=-2.0cm, xshift=-2.0cm] {CPU0};
\node (cpu1) [process, right of = cpu0, xshift=3.0cm] {CPU1};
\node (gpu0) [process2, below of = cpu0, xshift=-2.0cm, yshift=1.0cm] {Pascal P100 GPU};
\node (gpu1) [process2, right of = gpu0] {Pascal P100 GPU};
\draw [line] (in1) |- (cpu0);
\draw [line] (cpu0) -- node[anchor=south] {QPI}(cpu1);
\draw [line] (cpu0) -- node[anchor=east] {x16 PCIe}(gpu0);
\draw [line] (cpu0) -- (gpu1);
\node (gpu2) [process2, below of = cpu1, xshift=-2.0cm, yshift=1.0cm] {Pascal P100 GPU};
\node (gpu3) [process2, right of = gpu2] {Pascal P100 GPU};
\draw [line] (cpu1) -- (gpu2);
\draw [line] (cpu1) -- (gpu3);
\end{tikzpicture}
\caption{Hardware architecture layout of the Tiger cluster node.}
\label{fig:Hardware}
\end{center}
\end{figure*}
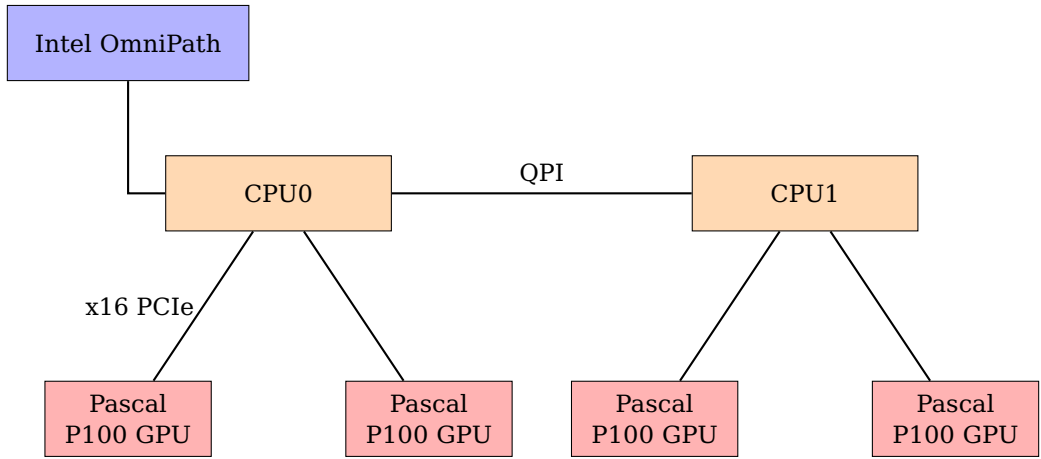

Scaling tests are performed on the Dell Linux cluster at Princeton University named "Tiger". It has a theoretical peak performance exceeding 27 petaflops, which is delivered by 320 NVIDIA Pascal P100 GPUs across 80 Intel Broadwell nodes. Each GPU is mounted on a dedicated x16 PCI express bus with 16 GB of HBM2 memory as shown in Fig.~\ref{fig:Hardware}.  The nodes of the cluster are interconnected by an Intel Omnipath high-speed interconnect. The CPUs are Intel Broadwell E5-2680v4 with 28 cores per node. 

\section{Neural Network architecture}

Recurrent Neural Networks (RNN) are a family of neural networks capable of processing sequential data of arbitrary length. Long short-term memory networks (LSTM) \cite{lstm}, a particularly successful form of RNNs, trained on large datasets can obtain remarkable performance results across a wide variety of domains -- from image captioning~\cite{mao2014deep}, to sentiment analysis and machine translation~\cite{Socher-etal:2013,NIPS2014_5346}. 

A LSTM unit uses no activation function within its recurrent components. Thus, the stored values are not iteratively expanded or squeezed over time, and the gradient does not tend to explode or vanish when trained. Instead, LSTMs contain gates controlling information flow, which are implemented using the logistic function.

\begin{figure*}
\begin{tikzpicture}[node distance=2cm]
\node (in1) [startstop] {Input: mini-batch of M time-series samples};
\node (lstm1) [process, below of = in1] {LSTM: (batch size, sequence length, feature dimension) = (256,128,9)};
\draw [arrow] (in1) -- (lstm1);
\node (lstm2) [process, below of = lstm1] {LSTM: (batch size, sequence length, num. hidden units in FC layer) = (256,128,200)};
\draw [arrow] (lstm1) -- (lstm2);
\node (fc2) [process, below of = lstm2] {FC layer: (batch size, sequence length, num. recurrent units in LSTM layer) = (256,128,200)};
\draw [arrow] (lstm2) -- (fc2);
\node (out1) [startstop, below of = fc2] {Output: (batch size, sequence length, num. hidden units in FC layer) = (256,128,1)};
\draw [arrow] (fc2) -- (out1);
\end{tikzpicture}
\caption{A typical neural network architecture for plasma disruption forecasting task.}
\label{fig:NNcartoon}
\end{figure*}
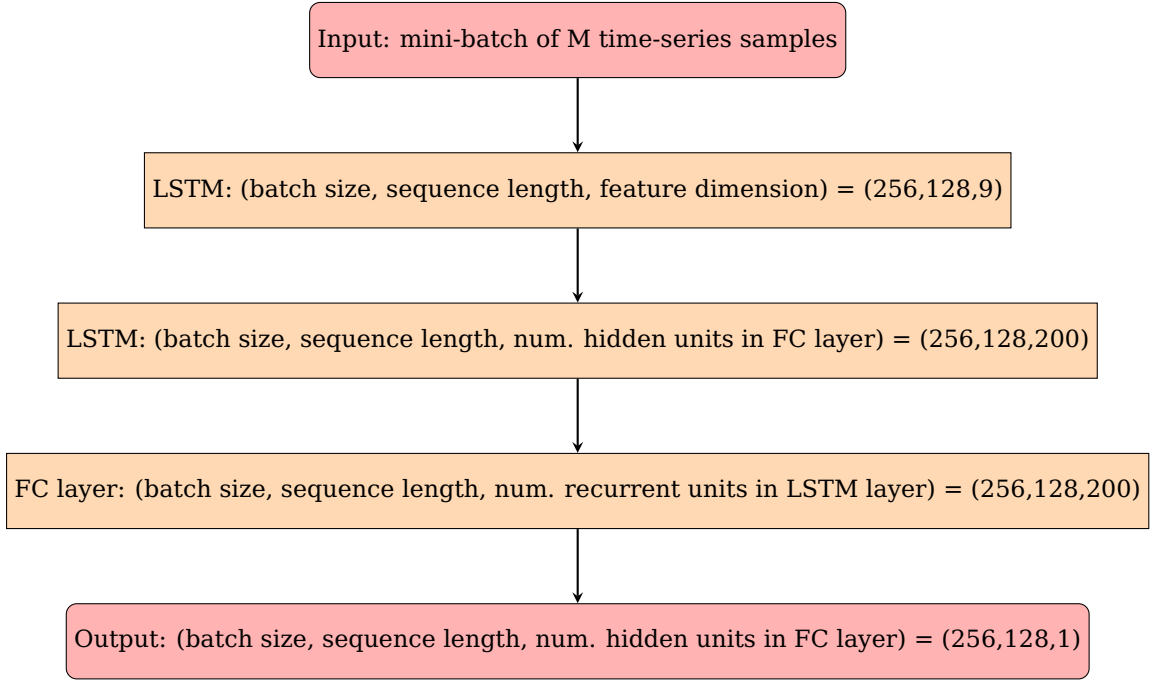

A typical neural network architecture for plasma disruption forecasting involves LSTMs and fully connected layers which are illustrated in Fig.~\ref{fig:NNcartoon}. 
Here, each LSTM layer has 200 hidden units, with application of L2 regularization and recurrent dropout. Fully connected layers with ReLU activation are applied for each temporal slice $t = 0...128$. Tunable parameters of neural network are determined via random search hyperparameter optimization. 

Neural networks are trained iteratively, making multiple passes over an entire data set before converging to a minimum. Each training iteration includes the forward propagation (\textbf{fprop}), loss calculation, backpropagation (\textbf{bprop}) followed by the weight update. Backpropagation through time (BPTT) is a gradient-based neural network training algorithm applied to train RNNs and LSTMs.

\section{Distributed computing framework}

We implement a distributed computing framework making use of TensorFlow for layer definitions, \textbf{fprop} and \textbf{bprop} steps, and a custom global weight update routine with parameter averaging implemented using CUDA-aware MPI to take advantage of high-speed network interconnects like OmniPath or Infiniband on HPC clusters, as well as the GPU-direct technology. 

The following summarizes the implementation of distributed data-parallel synchronous stochastic gradient descent with parameter averaging:
\begin{enumerate}
\item{Initialize the network parameters randomly based on the model configuration}
\item{\textit{broadcast} a copy of the current parameters to each worker}
\item{Perform \textbf{fprop} and \textbf{bprop} passes on each worker using a mini-batch $m_{i}$ of data}
\item{Aggregate gradients from each worker using \textit{allreduce}, then average them on worker with task $0$ to obtain global raw gradients}
\item{Update optimizer internal state and global weights using this global gradient on worker with task $0$}
\item{\textit{broadcast} global parameters after update back to all workers, return to the step (2) and repeat for mini-batch $m_{i+1}$ of data}
\end{enumerate}
Aggregating raw gradients allows using any optimizers in step (5), making the framework more flexible. In this paper, we focus on the stochastic gradient descent optimizer with momentum, which is given by the following update equations:
\begin{eqnarray}
\label{eq:sgd_mom}
H_{k} = m\cdot H_{k-1} - \lambda \Delta W, \\
W_{k} = W_{k-1} + H_{k}
\end{eqnarray}
With the above approach, the master has to collect all $N$ gradients in lock-step. An alternative, fault tolerant approach requires collecting a fraction of gradients (normally at 90-95\%) before proceeding to averaging, thus avoiding stalling the calculation in case of a slow node or a node failure. 

\section{Learning rate schedule}
\label{seq:LR}

The learning rate controlling magnitude of the weight update during stochastic gradient descent in Eq.~\ref{eq:sgd_mom} is lowered upon completion of each epoch. We use exponential learning rate schedule given by the following equation:
\begin{equation}
\lambda_{i} = \lambda_{0}\cdot \gamma^{i}
\end{equation}
where $\lambda_{0}$ is the base learning rate, $\gamma$ is the learning rate decay constant, and $i$ is the epoch number.

In a distributed regime, the learning rate is adjusted to facilitate reliable model convergence. First, the base learning rate is reduced as the number of workers $N$ increases:
\begin{equation}
\lambda_{0}(N,n) = \frac{\lambda_{0}}{1.0 + \frac{N}{n}}
\end{equation}
here, parameter $n$ controls the base learning rate adjustment, and is equal to the number of workers at which it is halved.

Secondly, the effective base learning rate, which is defined as a product of the number of workers and the base learning rate -- $\lambda_{0}\cdot N$ -- is clipped if it exceeds the maximum value of 0.1.

\begin{figure*}
\begin{center}
    \includegraphics[width=.70\textwidth]{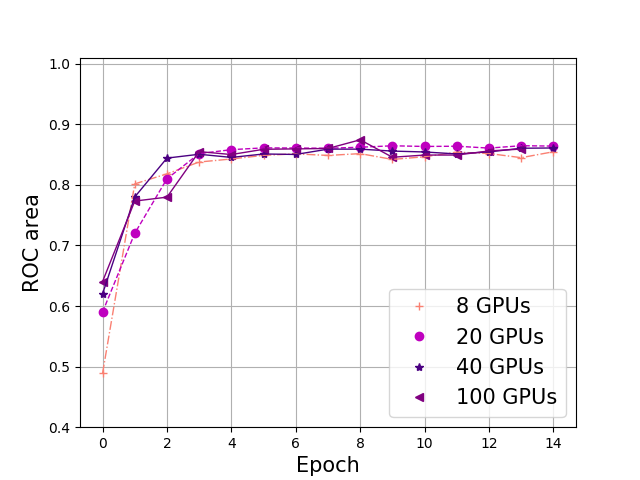}
    \caption{Validation level AUC per epoch calculated at single precision for 8, 20, 40 and 100 worker GPUs. Base learning rate: $\lambda_{0} = 0.0004$, batch size: $\beta_{0} = 256$. SGD optimizer with momentum, loss scaling factor: 10.0.}
    \label{dist_convergence} 
    \end{center}
\end{figure*}

Fig.~\ref{dist_convergence} shows validation level AUC per epoch calculated at single precision for up to 100 worker GPUs. While keeping batch size $\beta_{0}$ and base learning rate $\lambda_{0}$ the same in all cases, the number of workers has been varied. Effective batch size of the ensemble of workers is proportional to the number of workers: $\beta = N\cdot\beta_{0}$. Consequently, the model convergence may be affected for large ensemble sizes. As seen, the learning rate schedule introduced in Seq.~\ref{seq:LR} facilitates neural network convergence when training with up to 100 worker GPUs, achieving comparable AUCs at the plateau.

\section{Half-precision training modes}

Floating point formats consist of a sign, an exponent, and a mantissa. We use the single precision floating point format (\textbf{FP32}, 23-bit mantissa and 8-bit exponent) as a reference because this is the most widely used format in deep learning, especially for GPU computation. Half-precision floating point format (10-bit mantissa and 5-bit exponent) has a numerical range of (0.00006,65504). This narrow numerical range can potentially result in an overflow ("Inf/NaN" problem) or an underflow ("vanishing gradient") during training of neural networks. 

We show that the \textbf{FP16} with loss scaling has no significant impact on the neural network model convergence, resulting in an accuracy comparable to the \textbf{FP32} baseline, while allowing to train models with larger number of parameters, improving throughput and memory use.

To enable \textbf{FP16} training, a custom MPI data type of 2 contiguous bytes and a reduction operation (namely \textit{allreduce}) have been implemented. 

We distinguish the following precision modes:
\begin{enumerate}
\item{Math: matrix and element-wise multiplication during forward and backward passes}
\item{Synchronization: parameter averaging; precision of weights and gradients sent across network}
\item{Weight update}
\label{precisions}
\end{enumerate}
The baseline for the comparisons is the case when all -- math, synchronization and the weight update -- are of single precision.

The parts of the deep learning framework which are executed on CPU (including data preprocessing and normalization) all rely on the \textbf{FP32} and would not benefit from \textbf{FP16}, as the half type is currently emulated on CPU by converting the 32-bit floating point representations into a 16-bit floating point representation, thus leading to significantly slower runtimes.

\section{Performance evaluation}

The total time to process a mini-batch of data during synchronous SGD can be divided into computation $T_{batch}$ and synchronization $T_{sync}$ times. With the data-parallel implementation, computation time per mini-batch step remains constant in the number of workers, $T_{batch} \sim const$. Synchronization between workers is performed by means of a tree-like native MPI \textit{allreduce} operation, yielding logarithmic complexity $T_{sync} \sim log(N)$ --- providing a major benefit over the parameter server approach often used in distributed training. The amount of data processed during one mini-batch step increases linearly with the number of workers $N$. Thus, the number of mini-batches (and consequently the total time $T_{epoch}$) required for an epoch decreases linearly with $N$.
\begin{equation}
T_{epoch} \propto \frac{1}{N}(T_{batch} + T_{sync}) = \frac{1}{N}(A + B \cdot log(N)) = O(\frac{log(N)}{N})
\end{equation}



\begin{figure*}
\begin{center}
    \includegraphics[width=.70\textwidth]{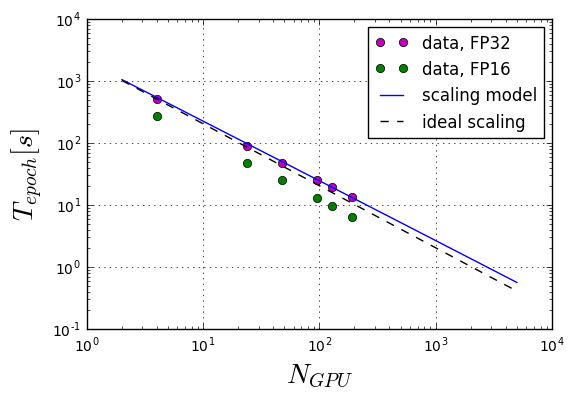}
    \caption{Compute time per epoch as a function of the number of worker GPUs for \textbf{FP16} and \textbf{FP32} at a maximum processing rate.}
    \label{tepoch} 
    \end{center}
\end{figure*}

Fig.~\ref{tepoch} shows the compute time per epoch as a function of the number of worker GPUs for \textbf{FP16} and \textbf{FP32} during distributed training on Pascal P100 GPUs at the maximum processing rate. 
As seen, the training algorithm provides linear runtime scaling as the number of worker GPUs $N$ increases.

The LSTM outputs a plasma disruptivity signal which is counted as an alarm when it passes a user-defined threshold. Calling an alarm at any point during a non-disruptive shot counts as a false positive (FP). Calling an alarm before the $30~$ms cutoff during a disruptive shot counts as a true positive (TP). The objective of optimization is to maximize true positives while minimizing false positives. Naturally, a higher threshold will lead to less alarms, and thus less false positives but also less true positives. Varying the threshold traces out an ROC curve that captures the full trade-off between TPs and FPs. The validation level area under the ROC curve, or AUC (area under curve), is the figure of merit we chose to characterize the quality of a given binary predictor, and applicability of the half-precision training.

\begin{figure*}
    \includegraphics[width=.49\textwidth]{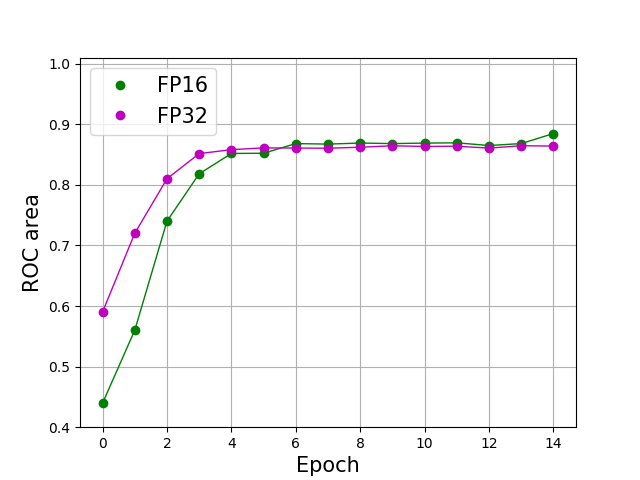}
    \includegraphics[width=.49\textwidth]{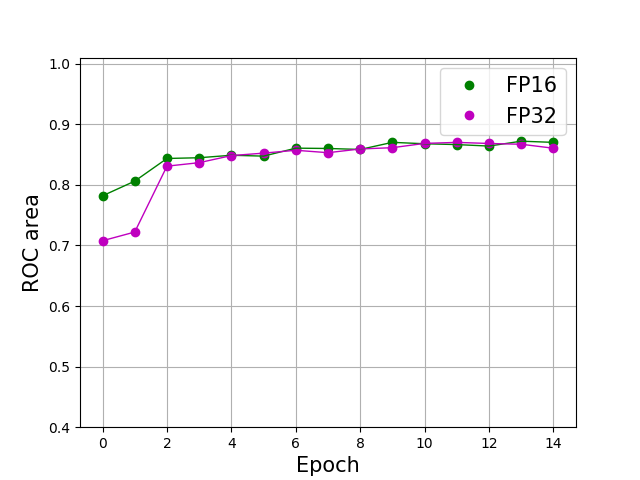}
    \caption{Validation level AUC per epoch calculated for \textbf{FP16} and \textbf{FP32} precisions. Left: base learning rate $\lambda_{0} = 0.0004$, right: base learning rate $\lambda_{0} = 0.001$. SGD optimizer with momentum, loss scaling factor: 10.0.}
    \label{val_rocs}
\end{figure*}

\begin{figure*}
\begin{center}
    \includegraphics[width=.75\textwidth]{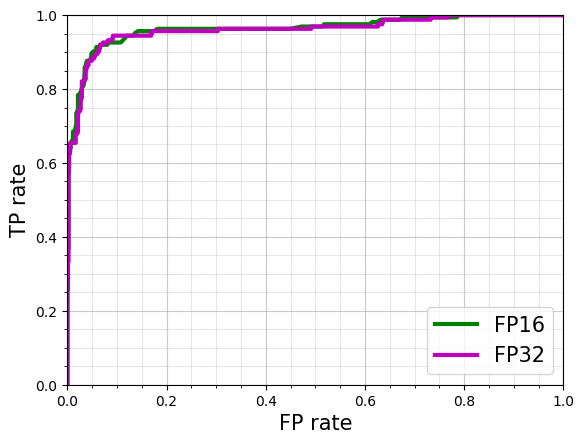}
    \caption{Test level ROC curve calculated for \textbf{FP16} and \textbf{FP32} precisions.} 
    \label{test_roc}
    \end{center}
\end{figure*}

As seen in Fig.~\ref{val_rocs}, both \textbf{FP16} and \textbf{FP32} show AUC as a function of epoch of similar shapes, reaching the plateau at around AUC=0.87 by the epoch 6. The test set AUC values are 0.96 $\pm$ 0.03 for both the \textbf{FP16} and \textbf{FP32} calculations, with the corresponding test ROC curves shown in Fig.~\ref{test_roc}. The test data set is obtained from a  different physical configuration leading to the difference between validation and test performances.

In a separate experiment, where neural network was trained till full convergence determined by "early stopping" with the validation AUC as a monitored quantity it took 8 and 9 passes over the training dataset to complete the training in the case of \textbf{FP16} and \textbf{FP32} respectively. 

The consistent accuracies between baseline and half-precision runs are achieved by applying a scalar multiplier $\alpha$ to the loss function before evaluating partial derivatives on the \textbf{bprop} step. For hinge loss function with $t$ denoting a classification label and $y$ the predicted neural network outcome we get:
\begin{equation}
L(y) = \alpha\cdot max(0,1-t\cdot y).
\end{equation}

\begin{figure*}
    \includegraphics[width=.49\textwidth]{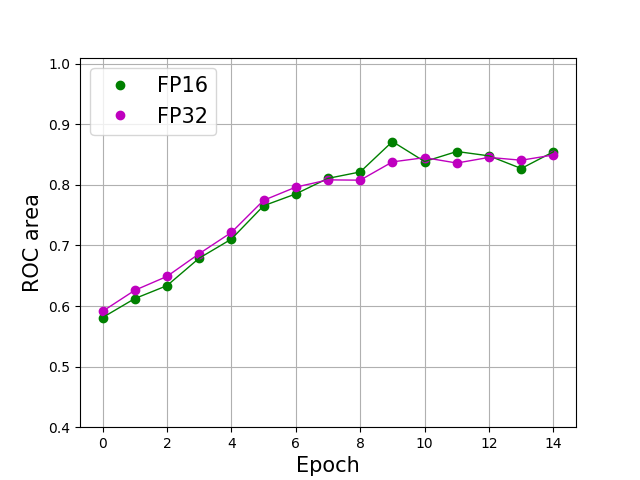}
    \includegraphics[width=.49\textwidth]{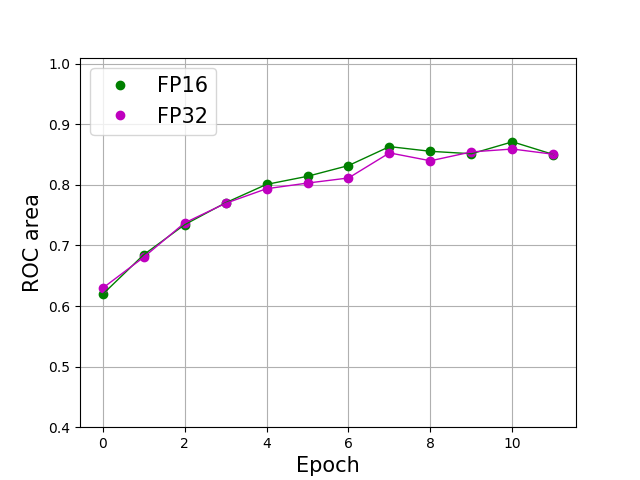}
    \caption{Validation level AUC per epoch calculated for \textbf{FP16} and \textbf{FP32} precisions for the Large Movie Review Dataset. Left: base learning rate $\lambda_{0} = 0.02$, right: base learning rate $\lambda_{0} = 0.05$.}
    \label{val_rocs_imdb}
\end{figure*}

The neural network training experiment is repeated on the benchmark IMDB dataset at different floating point precisions and learning rates.
As seen in Fig.~\ref{val_rocs_imdb}, both \textbf{FP16} and \textbf{FP32} show AUC as a function of epoch of similar shapes, reaching the plateau at around AUC=0.86 by the epoch 9 at $\lambda_{0} = 0.02$ and epoch 6 at $\lambda_{0} = 0.05$. 

\begin{figure*}
\begin{center}
    \includegraphics[width=.73\textwidth]{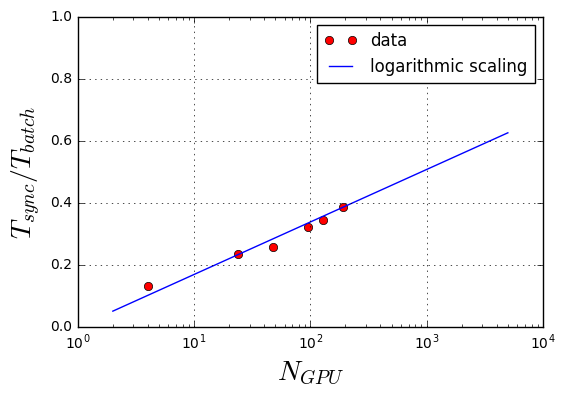}
    \caption{Ratio of synchronization to computation time time per mini-batch as a function of the number of worker GPUs.} 
    \label{synch}
    \end{center}
\end{figure*}

Fig.~\ref{synch} shows the ratio of synchronization time to computation time (each per mini-batch) as a function of the number of workers when training on Pascal P100 GPUs. 
As seen, communication time during the distributed training follows logarithmic scaling up to an $O(100)$ worker GPUs. 

Lower floating point precision allows training deeper models in a data-parallel fashion, and, potentially, use larger batch sizes. Tab.~\ref{tab:throughput} summarizes the maximum number of trainable parameters and number of layers for different floating point precisions and batch sizes. The model capacity is increased by stacking LSTM layers, while keeping the number of recurrent units and a sequence length fixed.
\begin{table}
\begin{center}
\begin{tabular}{|c|c|c|c|}
\hline
Precision & $N_{par}$ & $N_{layers}$ & Batch size \\
\hline
FP64 & $4.6\cdot 10^{6}$ & 15 & 256 \\
\hline
FP32 & $9.2\cdot 10^{6}$ &  29 & 256 \\
\hline
FP16 & $18.2 \cdot 10^{6}$  & 58 & 256 \\
\hline
FP64 &  $18.2 \cdot 10^{6}$ & 58 & 64 \\
\hline
FP32 & $36.3 \cdot 10^{6}$ &  118 & 64 \\
\hline
FP16 & $72.1 \cdot 10^{6}$ & 234 & 64 \\
\hline
\end{tabular}
\end{center}
\caption{Maximum capacity of the model in terms of number of trainable parameters, batch size, and equivalent model depth fitting in Pascal P100 GPU device memory for half-precision, single precision, and double floating point precision. A tuned configuration of the neural network for the plasma disruption forecasting task~\cite{Julian} consisting of 200 recurrent units and the sequence length of 128 is used.}
\label{tab:throughput}
\end{table}

The net gradient corresponding to the models in Tab.~\ref{tab:throughput} can be estimated as
$N_{par}\cdot$~Batch size~$\cdot$~Size of datatype, comprises 9.4~$GB/iter$. Using MPI \textit{allreduce} for synchronization, each GPU must send and receive about 9.4~$GB$ of data. Using the CUDA-aware MPI implementation (such as OpenMPI) allows data transfers between GPUs using GPUDirect random direct memory access with a bandwidth of roughly 10~$GB/s$, however, with the high-speed Intel Omnipath interconnect in our cluster the data can be transferred across the network with a bandwidth of 6.25~$GB/s$, which is slower. Since the limiting factor is the network communication, a single iteration requires about:
\begin{equation}
\frac{Net~gradient~size}{Bandwidth} = \frac{9.4~GB/iter}{6.25~GB/s} \approx 1500~ms/iter
\end{equation}

\section{Conclusions}

Training of deep recurrent neural networks with half-precision floats has been evaluated on a computing framework integrating TensorFlow with custom parameter averaging and global weight update routines implemented with CUDA-aware MPI. Tests with the JET plasma disruption time series dataset, and the benchmark Large Movie Review Dataset (IMDB) yield comparable validation level performances at the end of each epoch for half and single floating point precisions. 

A distributed data-parallel synchronous stochastic gradient descent approach showing strong nearly linear $O(\frac{log(N)}{N})$ runtime scaling is run on GPU clusters and evaluated. With half-precision runs, memory bandwidth and communication overhead is significantly reduced, allowing the fitting of larger models with over 70 million trainable parameters, and large batch sizes. 

A learning rate scheduling approach, reducing the base learning rate as a function of the number of workers followed by an exponential decay on a per-epoch basis is introduced to facilitate neural network convergence when training on HPC clusters with up to $O(100)$ worker GPUs.

Scalar multiplier $\alpha$ applied to the loss function before evaluating partial derivatives during the \textbf{bprop} step is found to be crucial for the model convergence at half-precision.

\appendix
\section{Datasets}
\subsection{JET}
\label{appendix:jet}
We present a summary of the JET dataset \cite{vega2013results} used throughout this paper. JET is the largest tokamak fusion experiment operating today and is situated in the UK. Plasma discharges (``shots'') range in length from $\sim 1$ to $\sim 40$ seconds and are sampled at a rate of $1~$ms. Thus, there are $O(10^3)$ to $O(10^4)$ timesteps per shot. Each shot consists of a scalar floating point value for each of the following measured plasma parameters for each timestep:
\begin{enumerate}
\item $q95$ plasma safety factor
\item $\beta$: plasma beta
\item $I_p$: plasma current
\item $l_i$: plasma internal inductance
\item $n$: plasma number density
\item $MLA$: amplitude of the locked mode signal
\item $P_{rad}$: radiated power
\item $E_{int}$: internal energy
\item $\frac{\partial E_{int}}{\partial t}$: time derivative of internal energy
\item $P_{in}$: input power
\end{enumerate}
A fraction of about $10\%$ of shots ends in a disruption. These shots are referred to as disruptive. All other shots are called non-disruptive. 

The dataset consists of $\sim 4300$ shots from JET experimental campaigns C15-C27b. During these campaigns the JET tokamak had carbon fiber composite (CFC) walls. These shots are used for training and validation in an $80/20$ split. The dataset also includes and $\sim 1100$ shots from the campaigns C28-30. In these more recent campaigns, the tokamak was upgraded to have a moder modern metallic wall. These shots are used for testing.

\subsection{Large Movie Review Dataset (IMDB)}

The dataset contains movie reviews along with their associated binary
sentiment polarity labels~\cite{imdb}. It often serves as a benchmark for
sentiment classification. It contains 50000 reviews split evenly between train
and test sets.  

In the entire collection, no more than 30 reviews are allowed for any
given movie because reviews for the same movie tend to have correlated
ratings. Further, the train and test sets contain a disjoint set of
movies.  In the labeled train and test sets, a negative review has a score of less or equal to 4 out of 10, and a positive review has a score of greater or equal to 7 out of 10. Thus reviews with more neutral ratings are not included in the train/test sets.

\begin{table}
\begin{center}
\begin{tabular}{|c|c|c|c|}
\hline
Signal & Test AUC & Train AUC & Val. AUC\\
\hline
Normalized Beta & 0.5340 & 0.5249& 0.4939\\
\hline
Locked mode amplitude & 0.7503& 0.7819& 0.7765\\
\hline
Input Power & 0.4158& 0.4527& 0.4538 \\
\hline
Radiated Power Core & 0.3597 &0.3732 & 0.3902 \\
\hline
Radiated Power Edge & 0.6122 & 0.5933& 0.6410 \\
\hline
q95 safety factor & 0.7914 & 0.7693& 0.7838 \\
\hline
Input Beam Torque & 0.5284 & 0.5406& 0.5242 \\
\hline
Plasma density & 0.5903 & 0.5817& 0.5763 \\
\hline
Electron dens. profile & 0.7008& 0.6976& 0.7647 \\
\hline
stored energy & 0.6389 & 0.6518& 0.6169\\
\hline
Electron temp. profile & 0.7142& 0.7226& 0.7648 \\ 
\hline
plasma current & 0.7143 & 0.7017& 0.7128 \\
\hline
plasma current direct. & 0.4787& 0.4940& 0.5221 \\
\hline
plasma current error & 0.4668 & 0.4652& 0.4674 \\
\hline
plasma current target & 0.7757& 0.7477& 0.7465 \\
\hline
internal inductance & 0.3609& 0.4026& 0.3741\\
\hline
\end{tabular}
\end{center}
\caption{Single signal summary for D3D dataset in terms of best validation level AUC, train and test level AUCs.}
\label{tab:throughput}
\end{table}

\begin{table}
\begin{center}
\begin{tabular}{|c|c|c|c|}
\hline
Signal & Test AUC & Train AUC & Val. AUC\\
\hline
Normalized Beta & 0.7805 & 0.8112 & 0.8074\\
\hline
Locked mode amplitude & 0.7785& 0.7492& 0.7799\\
\hline
Input Power & 0.7907& 0.7933& 0.7794\\
\hline
Radiated Power Core & 0.7903& 0.8311& 0.7764\\
\hline
Radiated Power Edge & 0.7891& 0.7244& 0.7794\\
\hline
q95 safety factor & 0.6777& 0.7228& 0.7676\\
\hline
Input Beam Torque & 0.7514& 0.7868& 0.7569\\
\hline
Plasma density & 0.7966& 0.7519& 0.7370\\
\hline
Electron dens. profile & 0.7976& 0.7732& 0.7716\\
\hline
stored energy & 0.7873& 0.7785& 0.7835\\
\hline
Electron temp. profile & 0.8069 & 0.7895& 0.7850\\
\hline
plasma current & 0.7993& 0.7974& 0.7829\\
\hline
plasma current direct. & 0.8034& 0.7731& 0.7813\\
\hline
plasma current error & 0.7439&  0.7596& 0.7742\\
\hline
plasma current target & 0.7802& 0.7991& 0.7710\\
\hline
internal inductance & 0.7461& 0.7640& 0.7694\\
\hline
\end{tabular}
\end{center}
\caption{Augmented study summary for D3D dataset in terms of best validation level AUC, train and test level AUCs. During training, augment one signal at random, one at a time. During inference, augment a fixed signal, one at a time.}
\label{tab:throughput}
\end{table}

\bibliographystyle{unsrt}  
\bibliography{references}  

\begin{thebibliography}{10}

\bibitem{imdb}
Andrew~L. Maas, Raymond~E. Daly, Peter~T. Pham, Dan Huang, Andrew~Y. Ng, and
  Christopher Potts.
\newblock Learning word vectors for sentiment analysis.
\newblock In {\em Proceedings of the 49th Annual Meeting of the Association for
  Computational Linguistics: Human Language Technologies}, pages 142--150,
  Portland, Oregon, USA, June 2011. Association for Computational Linguistics.

\bibitem{cpu}
Jeffrey Dean, Greg Corrado, Rajat Monga, Kai Chen, Matthieu Devin, Mark Mao,
  Marc\textquotesingle aurelio Ranzato, Andrew Senior, Paul Tucker, Ke~Yang,
  Quoc~V. Le, and Andrew~Y. Ng.
\newblock Large scale distributed deep networks.
\newblock In F.~Pereira, C.~J.~C. Burges, L.~Bottou, and K.~Q. Weinberger,
  editors, {\em Advances in Neural Information Processing Systems 25}, pages
  1223--1231. Curran Associates, Inc., 2012.

\bibitem{gpu}
Alex Krizhevsky, Ilya Sutskever, and Geoffrey~E Hinton.
\newblock Imagenet classification with deep convolutional neural networks.
\newblock In F.~Pereira, C.~J.~C. Burges, L.~Bottou, and K.~Q. Weinberger,
  editors, {\em Advances in Neural Information Processing Systems 25}, pages
  1097--1105. Curran Associates, Inc., 2012.

\bibitem{fpga}
Sang~Kyun Kim, Lawrence~C McAfee, Peter~Leonard McMahon, and Kunle Olukotun.
\newblock A highly scalable restricted boltzmann machine fpga implementation.
\newblock In {\em Field Programmable Logic and Applications, 2009. FPL 2009.
  International Conference on}, pages 367--372. IEEE, 2009.

\bibitem{DBLP:journals/corr/JozefowiczVSSW16}
Rafal J{\'{o}}zefowicz, Oriol Vinyals, Mike Schuster, Noam Shazeer, and Yonghui
  Wu.
\newblock Exploring the limits of language modeling.
\newblock {\em CoRR}, abs/1602.02410, 2016.

\bibitem{courbariaux+al-TR2014}
Matthieu Courbariaux, Yoshua Bengio, and Jean-Pierre David.
\newblock Training deep neural networks with low precision multiplications.
\newblock {\em arXiv e-prints}, abs/1412.7024, December 2014.

\bibitem{Lin:2016:FPQ:3045390.3045690}
Darryl~D. Lin, Sachin~S. Talathi, and V.~Sreekanth Annapureddy.
\newblock Fixed point quantization of deep convolutional networks.
\newblock In {\em Proceedings of the 33rd International Conference on
  International Conference on Machine Learning - Volume 48}, ICML'16, pages
  2849--2858. JMLR.org, 2016.

\bibitem{hubara+al-2016-quantized}
Itay Hubara, Matthieu Courbariaux, Daniel Soudry, Ran El-Yaniv, and Yoshua
  Bengio.
\newblock Quantized neural networks: Training neural networks with low
  precision weights and activations.
\newblock {\em arXiv e-prints}, abs/1609.07061, September 2016.

\bibitem{tensorflow2015-whitepaper}
Mart\'{\i}n Abadi, Ashish Agarwal, Paul Barham, Eugene Brevdo, Zhifeng Chen,
  Craig Citro, Greg~S. Corrado, Andy Davis, Jeffrey Dean, Matthieu Devin,
  Sanjay Ghemawat, Ian Goodfellow, Andrew Harp, Geoffrey Irving, Michael Isard,
  Yangqing Jia, Rafal Jozefowicz, Lukasz Kaiser, Manjunath Kudlur, Josh
  Levenberg, Dan Man\'{e}, Rajat Monga, Sherry Moore, Derek Murray, Chris Olah,
  Mike Schuster, Jonathon Shlens, Benoit Steiner, Ilya Sutskever, Kunal Talwar,
  Paul Tucker, Vincent Vanhoucke, Vijay Vasudevan, Fernanda Vi\'{e}gas, Oriol
  Vinyals, Pete Warden, Martin Wattenberg, Martin Wicke, Yuan Yu, and Xiaoqiang
  Zheng.
\newblock {TensorFlow}: Large-scale machine learning on heterogeneous systems,
  2015.
\newblock Software available from tensorflow.org.

\bibitem{vega2013results}
Jes{\'u}s Vega, Sebasti{\'a}n Dormido-Canto, Juan~M L{\'o}pez, Andrea Murari,
  Jes{\'u}s~M Ram{\'\i}rez, Ra{\'u}l Moreno, Mariano Ruiz, Diogo Alves, Robert
  Felton, JET-EFDA Contributors, et~al.
\newblock Results of the jet real-time disruption predictor in the iter-like
  wall campaigns.
\newblock {\em Fusion Engineering and Design}, 88(6):1228--1231, 2013.

\bibitem{Julian}
Julian Kates-Harbeck, Alexey Svyatkovskiy, Kyle Felker, Eliot Feibush, and
  William Tang.
\newblock Disruption forecasting in tokamak fusion plasmas using deep recurrent
  neural networks.
\newblock {\em Manuscript in preparation}, 2017.

\bibitem{lstm}
Felix~A Gers, J{\"u}rgen Schmidhuber, and Fred Cummins.
\newblock Learning to forget: Continual prediction with lstm.
\newblock 1999.

\bibitem{mao2014deep}
Junhua Mao, Wei Xu, Yi~Yang, Jiang Wang, Zhiheng Huang, and Alan Yuille.
\newblock Deep captioning with multimodal recurrent neural networks (m-rnn).
\newblock {\em ICLR}, 2015.

\bibitem{Socher-etal:2013}
Richard Socher, Alex Perelygin, Jean Wu, Jason Chuang, Christopher~D. Manning,
  Andrew~Y. Ng, and Christopher Potts.
\newblock Recursive deep models for semantic compositionality over a sentiment
  treebank.
\newblock In {\em Proceedings of the 2013 Conference on {E}mpirical {M}ethods
  in {N}atural {L}anguage {P}rocessing}, pages 1631--1642, Stroudsburg, PA,
  October 2013. Association for Computational Linguistics.

\bibitem{NIPS2014_5346}
Ilya Sutskever, Oriol Vinyals, and Quoc~V Le.
\newblock Sequence to sequence learning with neural networks.
\newblock In Z.~Ghahramani, M.~Welling, C.~Cortes, N.~D. Lawrence, and K.~Q.
  Weinberger, editors, {\em Advances in Neural Information Processing Systems
  27}, pages 3104--3112. Curran Associates, Inc., 2014.

\end{thebibliography}

\end{document}